\documentclass[letterpaper]{article} 
\usepackage{aaai2026}  
\usepackage{times}  
\usepackage{helvet}  
\usepackage{courier}  
\usepackage[hyphens]{url}  
\usepackage{graphicx} 
\usepackage{natbib}  
\usepackage{caption} 
\usepackage{algorithm}
\usepackage{algorithmic}
\usepackage{booktabs}
\usepackage{multirow}
\usepackage[dvipsnames]{xcolor}
\usepackage{amsmath}
\usepackage{amssymb}
\UseRawInputEncoding
\usepackage{newfloat}
\usepackage{listings}
\DeclareCaptionStyle{ruled}{labelfont=normalfont,labelsep=colon,strut=off} 
\floatstyle{ruled}
\newfloat{listing}{tb}{lst}{}
\floatname{listing}{Listing}
\title{Modality-Specific Enhancement and Complementary Fusion for Semi-Supervised Multi-Modal Brain Tumor Segmentation}

\author{
    Tien-Dat Chung\textsuperscript{\rm 1}\equalcontrib,
    Ba-Thinh Lam\textsuperscript{\rm 4}\equalcontrib,
    Thanh-Huy Nguyen\textsuperscript{\rm 2},
    Thien Nguyen\textsuperscript{\rm 2},
    Nguyen Lan Vi Vu\textsuperscript{\rm 3},
    Hoang-Loc Cao\textsuperscript{\rm 1},
    Phat K. Huynh\textsuperscript{\rm 1},
    Min Xu\textsuperscript{\rm 2},
}
\affiliations{

    \textsuperscript{\rm 1} PASSIO Lab, North Carolina A\&T State University, USA \\
    \textsuperscript{\rm 2} Carnegie Mellon University, USA \\
    \textsuperscript{\rm 3} Ho Chi Minh University of Technology, Vietnam \\
    \textsuperscript{\rm 4} University of North Carolina at Charlotte, USA \\
}

\usepackage{bibentry}

\usepackage{svg}
\begin{document}

\maketitle

\begin{abstract}
Semi-supervised learning (SSL) has become a promising direction for medical image segmentation, enabling models to learn from limited labeled data alongside abundant unlabeled samples. However, existing SSL approaches for multi-modal medical imaging often struggle to exploit the complementary information between modalities due to semantic discrepancies and misalignment across MRI sequences. To address this, we propose a novel semi-supervised multi-modal framework that explicitly enhances modality-specific representations and facilitates adaptive cross-modal information fusion. Specifically, we introduce a Modality-specific Enhancing Module (MEM) to strengthen semantic cues unique to each modality via channel-wise attention, and a learnable Complementary Information Fusion (CIF) module to adaptively exchange complementary knowledge between modalities. The overall framework is optimized using a hybrid objective combining supervised segmentation loss and cross-modal consistency regularization on unlabeled data. Extensive experiments on the BraTS 2019 (HGG subset) demonstrate that our method consistently outperforms strong semi-supervised and multi-modal baselines under 1\%, 5\%, and 10\% labeled data settings, achieving significant improvements in both Dice and Sensitivity scores. Ablation studies further confirm the complementary effects of our proposed MEM and CIF in bridging cross-modality discrepancies and improving segmentation robustness under scarce supervision.
\end{abstract}


\section{Introduction}

\begin{figure}[t]
    \centering
    \includegraphics[width=\linewidth]{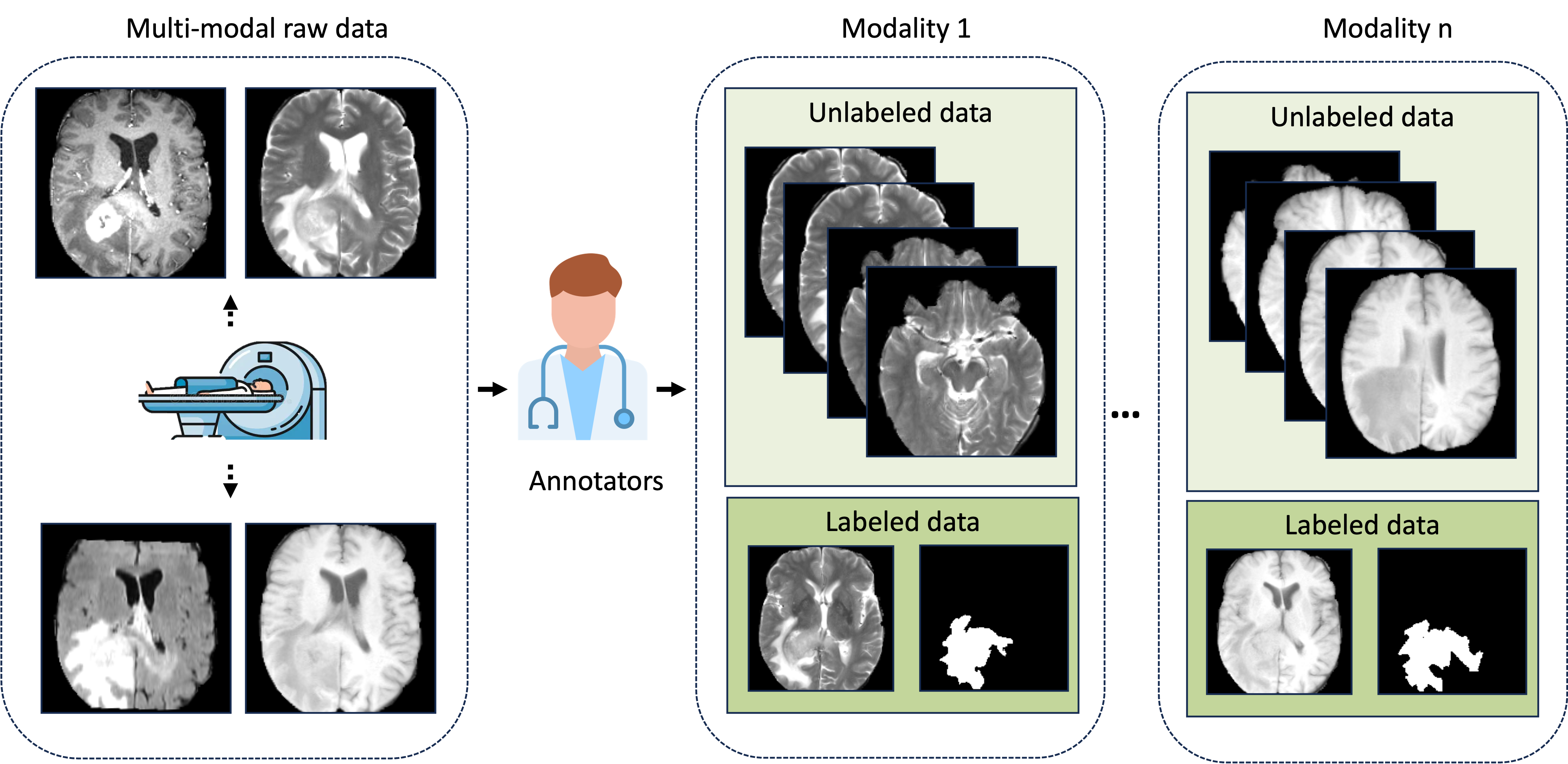}
    \caption {Our semi-supervised cross-modality data settings is illustrated. Specifically, the raw MRI sequences obtained from the same patient which subsequently partly annotated by radiologists in order to produce a multi-modal semi-supervised dataset.}
    \label{fig:abtract}
\end{figure}
Brain tumours are essentially one of the most life-threatening neurological diseases despite the rise of surgical techniques and modern therapies. Magnetic Resonance Imaging (MRI) plays a vital role in brain disease diagnosis due to its ability to show soft-tissue contrast. In clinical practice, radiologists often utilize MRI sequences, namely T1, T1CE, T2, and FLAIR to carefully consider the different anatomical structures of the tumour. Each sequence captures distinct aspects of the tumour, which are complementary to each other. Thereby, T1 provides anatomical context, T1CE highlights contrast-enhancing tumour regions, and FLAIR carries peritumoral oedema \cite{brast_intro}. By jointly analyzing these sequences, clinicians have the ability to produce a comprehensive diagnosis of the tumour. Therefore, the field of multi-modal learning has surged recently, which has shown promising results for brain tumour segmentation \cite{ma, sup_2, sup_3}. They integrate additional information from multiple MRI modalities by developing multi-modal networks that can capture semantic representations of tumour within different modalities and fuse them into a rich representation for later decoding. 

However, the success of multi-modal learning goes along with a major bottleneck: it is inherently data-hungry. In order to achieve high performance requires large-scale datasets where all modalities are meticulously annotated or registered. In clinical practice, acquiring such labeled multi-modal datasets is challenging and costly. To mitigate the reliance on extensive annotated data, semi-supervised learning (SSL) methods have emerged, allowing models to effectively leverage a large amount of unlabeled data while training along with labeled data. For instance,  Tarvainen et al. \cite{meanteacher} introduce a teacher-student consistency framework that enforces stable predictions between the two models, while Lee et al. \cite {pseudo} utilize confident model predictions as pseudo-labels for unlabeled data. Furthermore, a cross-model learning scheme in which two networks exchange pseudo-labels to mutually guide each other was proposed by \cite {cps}. 

As a result, several SSL approaches tailored for multi-modal medical image segmentation have been proposed. For example, Zhou et al. utilized Cross Modality Collaboration (CMC) \cite {cmc}, Zhang et al. introduced Multi-modal contrastive mutual learning \cite {cml}, or a multi-stage fusion of \cite{complex} have successfully employed consistency regularization, mutual learning, and pseudo-labeling to fuse information and improve robustness under data scarcity. 

Despite the notable results of these methods, effectively diminishing the inherent discrepancy and potential misalignment between modalities when using consistency constraints or utilizing pseudo-labels is still a common challenge. This observation motivates our work, such that we develop a semi-supervised multi-modal framework that focuses on enhancing the intrinsic feature within each modality, facilitating the feature fusion stage for complementary information exchange. 


As shown in Figure~\ref{fig:abtract}, our data setting has the ability of training a semi-supervised cross-modality model which uniquely enables complementary information exchange between modalities using both labeled and unlabeled data, bridging the gap in the single-modality SSL setting. 

Our main contributions of this work are: 
\begin{itemize}
    \item We propose a Modality-specific Enhancing Module (MEM) that extracts modality-specific semantic information via channel-wise attention while preserving the original knowledge of the input modality. 
    \item We introduce an adaptively learnable Complementary Information Fusion (CIF) that serves as a bridge for two modalities to exchange complementary information. 
    \item We employ a semi-supervised training framework that leverages both labeled and unlabeled data with supervised segmentation loss and cross-modal consistency loss, respectively. Additionally, extensive experiments on BraTS 2019 dataset are conducted to show the effectiveness of proposed modules compared to previous works.
\end{itemize}
\section{Related Works}
\subsection{Semi-supervised Medical Image Segmentation}

Semi-supervised learning (SSL) has emerged as a compelling paradigm for medical image segmentation, aiming to reduce the dependence on large annotated datasets by leveraging abundant unlabeled data alongside a limited number of labeled samples.  
Early SSL frameworks can be grouped into several main categories: \textit{self-training}~\cite{bai2017semi, nguyen2025duetmatch, nguyen2025semi, vu2025semi}, which iteratively refines pseudo-labels; \textit{adversarial learning}~\cite{li2020shape}, which enforces distribution-level consistency via discriminator networks; \textit{co-training}~\cite{zhou2019deep,peng2020self,wang2021self, nguyensemi}; and \textit{consistency regularization}~\cite{bortsova2019semi,tarvainen2017mean, nguyen2025adaptive}, which constrains model predictions to remain invariant under data perturbations.  
Among these, consistency-based frameworks, especially the Mean Teacher~\cite{tarvainen2017mean} and uncertainty-aware consistency schemes~\cite{yu2019uncertainty,li2020transformation}, have achieved state-of-the-art results by promoting stable training under sparse supervision.

However, the majority of SSL segmentation methods are designed for \textit{single-modality} images (e.g., CT or MRI), overlooking the complementary anatomical and appearance information available in multimodal clinical imaging.  
Extending SSL to the multimodal domain introduces new challenges: modality misalignment, varying intensity distributions, and inconsistent anatomical contrast. Thus, recent works have begun to explore multimodal semi-supervised approaches that unify consistency regularization with cross-modality collaboration to better exploit the latent relationships between modalities~\cite{cml}.

\subsection{Multi-modal Semi-supervised Medical Image Analysis}

While multimodal learning enriches representational power, its reliance on fully labeled datasets limits practical deployment. Semi-supervised learning (SSL) offers a compelling solution by leveraging unlabeled multimodal data alongside limited annotations.  Early multimodal SSL efforts, such as DAFNet~\cite{chartsias2020dafnet} and few-shot multimodal segmentation frameworks~\cite{mondal2018few}, introduced feature disentanglement and generative alignment \cite{nguyen2023towards, truong2023delving, nguyen5109180mv}, but their tightly coupled architectures required all modalities during inference, restricting flexibility in clinical practice.

To address this limitation, Zhang \textit{et al.}~\cite{cml} proposed the \textit{Semi-supervised Contrastive Mutual Learning (Semi-CML)} framework, which performs low-coupling cross-modality learning using an \textit{Area-Similarity Contrastive (ASC)} loss. This loss enhances mutual consistency between modalities by integrating Dice similarity into contrastive learning, thereby capturing area-context information crucial for segmentation.  
Furthermore, a \textit{Pseudo-label Re-learning (PReL)} scheme, guided by a \textit{Best-model Moving Average (BMA)} teacher, mitigates modality performance gaps by transferring reliable pseudo-labels from the stronger modality to the weaker one.

Building on this, Zhou \textit{et al.}~\cite{cmc} introduced the \textit{Cross-Modality Collaboration (CMC)} framework, which distills modality-independent knowledge via a \textit{Channel-wise Semantic Consistency (CSC)} loss and enforces anatomical alignment through \textit{Contrastive Consistent Learning}.  
Unlike previous multimodal SSL methods, CMC explicitly regularizes feature alignment and structural consistency across modalities, enabling robustness under scarce labels and misaligned multimodal conditions.

Together, these advances demonstrate that unifying cross-modality collaboration with semi-supervised consistency regularization offers a promising paradigm for clinically deployable multimodal medical image segmentation.

\section{Methodology}

\begin{figure*}[!ht]
    \centering
    \includegraphics[width=\linewidth]{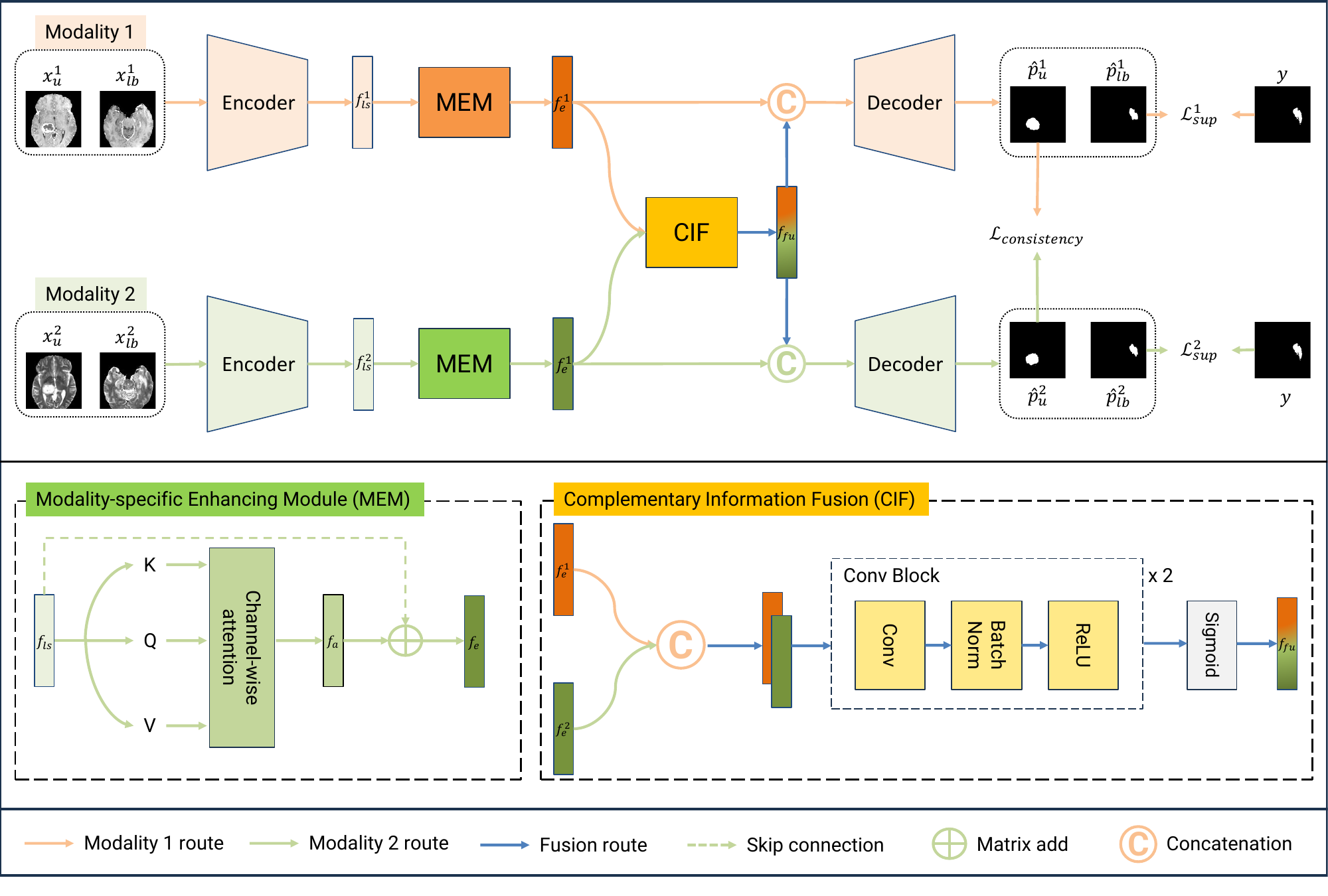}
    \caption{Illustration shows all components in our proposed framework, which consists of two branches according to two different input modalities. Each branch uses the same architecture and process, thereby a distinctive feature extracted by a U-Net encoder is processed by channel-wise attention of a Modality-specific Enhancing Module for enhancing modality-specific knowledge, facilitating the next feature fusion stage to produce a comprehensive feature representation at a Complementary Information Fusion layer. This representation finally concatenates with enhanced features of each branch, subsequently passing a U-Net decoder. The training procedure is jointly optimized by a semi-supervised strategy composed of supervised and consistency loss.}
    \label{fig:framework}
\end{figure*}

Our proposed framework for semi-supervised multi-modal information complementarity learning is illustrated in Fig. \ref{fig:framework}. Two input modalities are first processed by a dual-branch segmentation network with distinct feature extractions, followed by our Modality-specific Enhancing Modules to enhance modality information of early extracted features by adding the inherent semantic knowledge that exclusively belongs to each modality to its original. This procedure is intended to prime it for a better fusion feature stage. Subsequently, an adaptively learnable Complementary Information Fusion (CIF) is employed to combine the enhanced features from two modalities for a more comprehensive representation, which is concatenated back to the enhanced embedding of each branch to play as a supplementary perspective from the other modality. Two decoders are then used to generate segmentation masks. Notably, we optimized the model by supervised loss for labeled data and a consistency learning strategy for unlabeled data.





\subsection{Modality-specific Enhancing Module}
We denoted the multi-modality dataset \(\mathbf{D}_{lb}\) and \(\mathbf{D}_{u}\) as: 
\begin{equation}
\begin{split}
    \mathbf{D}_{lb} &= \{\{(x^1_{lb,i}, y_{i}) \}^N_{i=1}, \{(x^2_{lb,i}, y_{i}) \}^N_{i=1}\},  \\
    \mathbf{D}_{u} &= \{(x^1_{u,j})^M_{j=1}, \{(x^2_{u,j})\}^M_{j=1}\},
\end{split}
\end{equation}
where \(x_{lb,i}\) and \(x_{u,j}\) indicate the labeled and unlabeled data in modality 1 and 2, respectively. Both \(x^1_{lb,i}\) and \(x^2_{lb,i}\) have the same segmentation mask \(y_{i}\). The N and M denote the number of labeled and unlabeled samples, and \(N \ll M\).

In clinical diagnosis, radiologists often utilize multiple modalities to obtain a better result from patients'  MR images. We gained insight into such behavior, different modalities provide complementary perspectives as modality-specific knowledge, e.g., T1 for anatomy and T2 for fluid, containing the information genuinely important for the segmentation result and solely associated with that modality. In the deep learning realm, the self-attention mechanism in computer vision is tailored to point out the relationship among different areas within an image globally, and such an unmatched mechanism effectively highlights the unique connection in semantic meaning among multiple areas in a particular image. As a result, we designed a channel-wise modality-specific attention module that extracts modality-specific information in the feature from the last convolutional layers of the U-Net encoder. Our module is basically intended not only to enhance the modality-specific knowledge but also to preserve its original knowledge, which is the inherent semantic meaning extracted from the powerful U-Net encoder at the early stage. 

Each modality \(x^{1}\) and \(x^{2}\) is fed into its respective encoder; notably, the encoders are the same in architecture, but we use two distinct encoders for different modalities for effective modality-specific features extraction. Each encoder consists of four maximum poolings in order to reduce the original image resolution by 16 times, and it includes five layers of convolutional blocks, thus producing five different feature maps with corresponding dimensions of 32, 64, 128, 256, and 512. We, however, solely use the last feature map \(f^{1}_{ls}\) and \(f^{1}_{ls}\) from both branches, which is a 512-dimensional feature, because of its inherent abundant semantic information compared to the others. To effectively capture the intrinsic modality-specific feature from \(f^{1}_{ls}\) and \(f^{1}_{ls}\), we introduce a channel-wise Modality-specific Enhancing Module (MEM) which focuses on emphasizing the correlations between channels in the feature map. 

MEM of two different branches leverages its corresponding \(f_{ls}\) to produce an enhanced feature map which contains both original information and exclusive modality-specific information. Specifically, we first generate a channel-aware feature map \(f_{a}\) from each \(f_{ls}\), which carries the intrinsic modality-specific knowledge by a channel-wise self-attention mechanism. Then, to ensure the preservation of information within \(f_{ls}\), we employ a skip connection that brings the original feature representation towards, resulting in an enhanced feature \(f_{e}\). Finally, we obtain two enhanced features of two modalities, facilitating a better fusion stage.


\subsection{Complementary Information Fusion Module}

In order to fuse the complementary information from different modalities, we designed an adaptively learnable fusion layer module to combine high-dimensional modality-specific features \(f_{e}\) from two branches of modality. 

The two features are first concatenated by the fusion layer along the dimension layer, then the joint representation is passed through two conv blocks consisting of a convolutional layer, batchnorm and activation. These conv blocks are responsible for adaptively learn the extent of each modality to be combined into a high-quality fused representation \(f_{fu}\). Subsequently, the sigmoid activation is applied to ensure stability while training. Finally, the fused representation is concatenated back to the modality-specific feature \(f_{e}^{1}\) and \(f_{e}^{2}\) from each branch for the model to exchange complementary information from the other. After the concatenation, we obtain an updated features containing modality-specific knowledge from both input modalities, which is now a new version of the last convolutional layer's feature map to be fed along with the other feature maps of encoders (e.g., 32-, 64-, 128-, and 256-dimensional feature maps) into respective decoders for a segmentation prediction.

\subsection{Semi-supervised cross-modality training}
During training, our model is optimized using supervised loss on labeled data, which is defined as below: 

\begin{equation}
    \mathcal{L}_{sup} = \beta \mathcal{L}_{CE}(\hat{p}_{lb},y) + \gamma \mathcal{L}_{DICE}(\hat{p}_{lb},y),
\end{equation}
with \(\mathcal{L}_{CE}\) is cross-entropy, \(\mathcal{L}_{DICE}\) is dice loss, predictions \(\hat{p}_{lb}\) and annotation \(y\). The coefficient \(\beta \) and \(\gamma\) to weight the contribution of losses. Losses from both modalities are summed:
\begin{equation}
    \mathcal{L}^{total}_{sup} = \mathcal{L}^1_{sup} + \mathcal{L}^2_{sup}.
\end{equation}

In addition, to exploit the redundant amount of unlabeled data, we introduce a consistency regularization to enforce the model to generate consistency in its predictions that rely on cross-modal mutual knowledge. According to this operation, two different models can learn from each other by leveraging the complementary information between modalities. Therefore, we design a consistency loss function based on MSE, enabling the ability of learning the mutual information by minimizing the differences in pixel-level predictions on two different modalities, which is defined as below:

\begin{equation}
    \mathcal{L}_{consistency} = \frac{1}{M}\sum\|\hat{p}^1_u - \hat{p}^2_u \|^2
\end{equation}

where \(\hat{p}^1_u\) and \(\hat{p}^2_u\) are the prediction masks of unlabeled data of different modalities. Thereby, the final loss function is defined as below:

\begin{equation}
     \mathcal{L}_{final} = \mathcal{L}^{total}_{sup} + \gamma \mathcal{L}_{consistency}
\end{equation}
We empirically found the optimal value for \(\gamma\) is \(0.01\). 

\section{Experiments}

\subsection{Dataset \& Metric}
\paragraph{Datasets}

We conducted extensive experiments on a public dataset, the BraTs 2019 Challenge, which contains 259 high-grade gliomas (HGG) and 76 low-grade gliomas (LGG) data, but we only used the HGG data in our experiments. The dataset contains four modalities: T1, T1CE, T2, and FLAIR. The preprocessing procedure includes division of the dataset into a training set and a test set with 80\% and 20\%, respectively, slicing 3D image volumes into 2D images and removing the slices without lesions to prevent data imbalance. As a result, we obtain 13,598 slices, which are normalized to zero mean and unit variance, subsequently center-cropped to the size of 160 \(\times\) 160.
\begin{figure*}[!h]
    \centering
    \includegraphics[width=\linewidth]{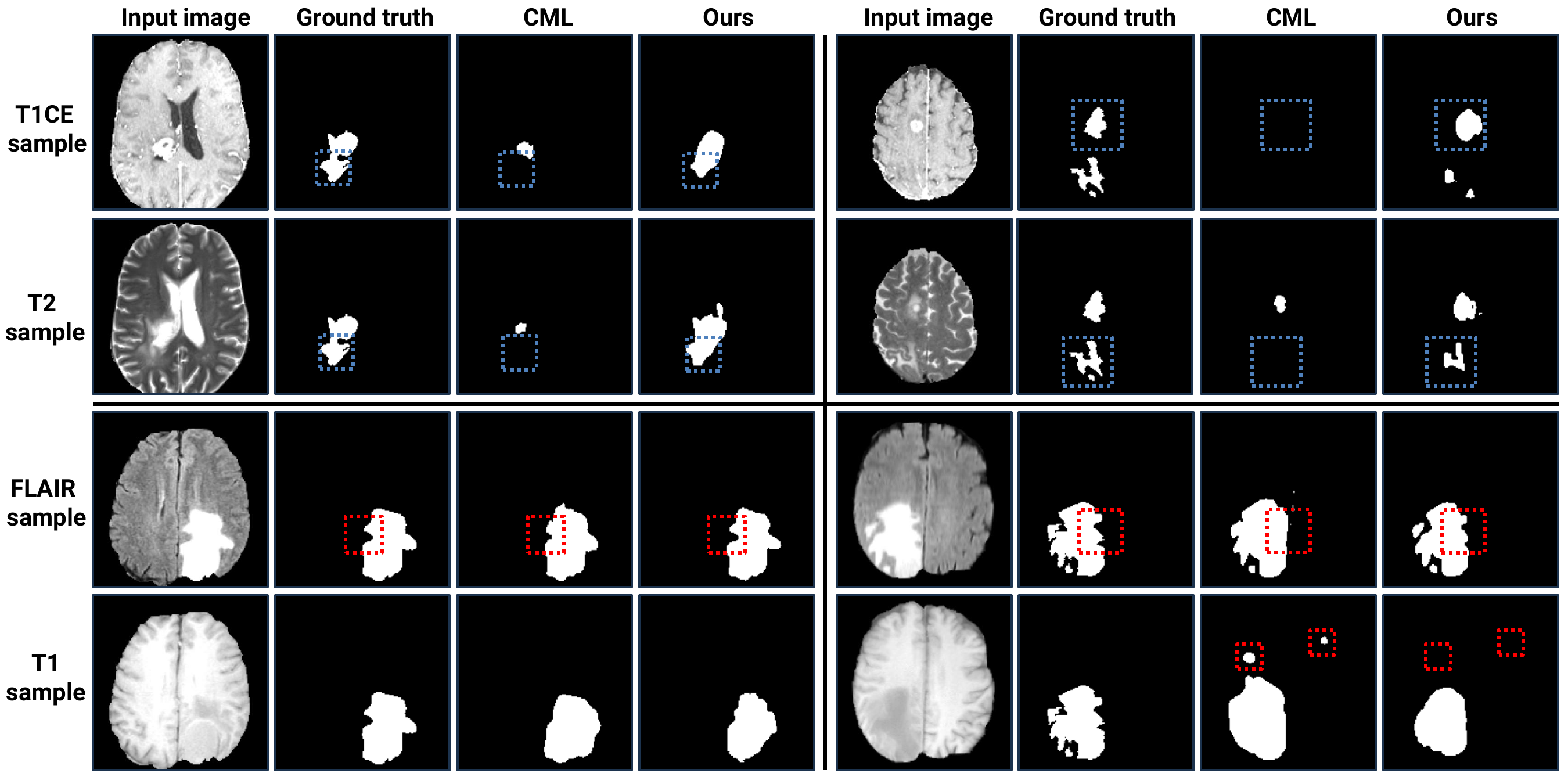}
    \caption{Visualization compares our method and CML, both models trained on T2-T1CE and T1-FLAIR modality combination with 10\% of labeled data. The figure includes eight samples of four modalities, such that each row shows two samples of a specific modality. Note that \textcolor{Cerulean}{blue square} boxes indicate a higher recall prediction, \textcolor{red}{red square} boxes are for a higher precision prediction in comparison between our model and CML.}
    \label{fig:vis}
\end{figure*}

\paragraph{Metrics}
In this study, we primarily employed two metrics: Dice Score Coefficient (DSC) and Sensitivity (Sens). The dice score measures the overlap area between ground truth and predicted segmentation; a higher DSC indicates a closer segmentation result of our model compared to the ground truth. The sensitivity, also known as recall, quantifies the extent to which the model can correctly identify the positive pixels among all actual positive pixels. These two metrics are widely used in medical image segmentation; thereby dice score averages the overall segmentation performance, while the sensitivity presents the model's capability to minimize the false negatives, which is significantly essential in medical image analysis.
\subsection{Settings}
In order to ensure a fair comparison, we employed all experiments on the same Pytorch version, run on RTX 4060TI GPU (16GB), fixing a random seed for both the training and evaluation process. We randomly select a proportion of 1\%, 5\%, and 10\% of the entire dataset as labeled data, while the rest is an unlabeled counterpart for our semi-supervised training framework. The Adam optimizer is used to optimize our experiments with a base learning rate at \(6 \times 10^{-3}\) and a weight decay of \(4\times  10^{-4}\). In addition, we set the coefficient \(\gamma\) and \(\beta\) for dice coefficient loss and cross entropy, both are \(1.0\). During evaluation, we average dice scores and sensitivity among patients' slices of 3D image volumes.
\subsection{Results}
\begin{table*}[t]
\centering
\caption{Segmentation results of our method and multi-modal semi-supervised methods on \textbf{BraTS (T1--FLAIR)} dataset with 1\%, 5\%, and 10\% labeled data in terms of DSC and Sens.}
\label{tab:t1flair}
\resizebox{0.8\textwidth}{!}{%
\begin{tabular}{llcccccc}
\toprule
\multirow{2}{*}{Methods} & \multirow{2}{*}{Inference modality} &
\multicolumn{2}{c}{1\% labeled} & \multicolumn{2}{c}{5\% labeled} &
\multicolumn{2}{c}{10\% labeled} \\
\cmidrule(lr){3-4}\cmidrule(lr){5-6}\cmidrule(lr){7-8}
 &  & DSC & Sens & DSC & Sens & DSC & Sens \\
\midrule
Mean Teacher \textcolor{gray}{(NeurISP'17)}   & T1--FLAIR & 0.3929 & 0.4132 & 0.4433 & 0.4582 & 0.4912 & 0.5073 \\
        & FLAIR     & 0.2407 & 0.2166 & 0.2178 & 0.2071 & 0.2946 & 0.3142 \\
\hline
DTC \textcolor{gray}{(AAAI'21)}  & T1--FLAIR & 0.3712 & 0.3825 & 0.4319 & 0.4515 & 0.5016 & 0.5174 \\
        & FLAIR     & 0.2042 & 0.1827 & 0.2248 & 0.2671 & 0.3023 & 0.3380 \\
\hline
DTML \textcolor{gray}{(PRCV'21)} & T1--FLAIR & 0.3926 & 0.3962 & 0.4306 & 0.4602 & 0.5080 & 0.5149 \\
        & FLAIR     & 0.2459 & 0.2209 & 0.2466 & 0.2465 & 0.2809 & 0.3296 \\
\hline
SASS \textcolor{gray}{(MICCAI'20)} & T1--FLAIR & 0.3868 & 0.4081 & 0.4232 & 0.4416 & 0.4939 & 0.5216 \\
        & FLAIR     & 0.2744 & 0.2387 & 0.1932 & 0.2407 & 0.2667 & 0.4287 \\
\hline
UAMT \textcolor{gray}{(MICCAI'19)} & T1--FLAIR & 0.3956 & 0.4195 & 0.4615 & 0.5733 & 0.5129 & 0.5891 \\
        & FLAIR     & 0.2601 & 0.2275 & 0.3048 & 0.3087 & 0.4047 & 0.4895 \\
\hline
UMCT \textcolor{gray}{(MedIA'20)}& T1--FLAIR & 0.4054 & 0.4243 & 0.4525 & 0.5602 & 0.5040 & 0.5230 \\
        & FLAIR     & 0.3372 & 0.3365 & 0.2610 & 0.2662 & 0.3089 & 0.3083 \\
\hline
SPCT \textcolor{gray}{(MedIA'21} & T1--FLAIR & 0.4002 & 0.4220 & 0.4638 & 0.4860 & 0.5131 & 0.5355 \\
        & FLAIR     & 0.2595 & 0.2357 & 0.2444 & 0.2393 & 0.2934 & 0.3684 \\
\hline
DAFNet \textcolor{gray}{(T-MI'20)}  & T1--FLAIR & 0.4100 & 0.5900 & 0.5050 & 0.5300 & 0.5260 & 0.5880 \\
        & FLAIR     & 0.3460 & 0.4960 & 0.3720 & 0.4710 & 0.4540 & 0.5660 \\
        & FLAIR     & 0.3173 & 0.4565 & 0.2207 & 0.1968 & 0.2575 & 0.2544 \\
\hline
CML \textcolor{gray}{(MedIA'23)}    & FLAIR     & 0.4337 & 0.4833 & 0.4854 & 0.5858 & 0.5302 & 0.6079 \\
\hline
Ours    & T1--FLAIR & \textbf{0.7232}& \textbf{0.6699} & \textbf{0.7728} & \textbf{0.8214} & \textbf{0.7970} & \textbf{0.8579} \\
\bottomrule
\end{tabular}}
\end{table*}

\begin{table*}[!h]
\centering
\caption{Segmentation results of our method and multi-modal semi-supervised methods on \textbf{BraTS (T2--T1CE)} dataset with 1\%, 5\%, and 10\% labeled data in terms of DSC and Sens.}
\label{tab:t2t1ce}
\resizebox{0.8\textwidth}{!}{%
\begin{tabular}{llcccccc}
\toprule
\multirow{2}{*}{Methods} & \multirow{2}{*}{Inference modality} &
\multicolumn{2}{c}{1\% labeled} & \multicolumn{2}{c}{5\% labeled} &
\multicolumn{2}{c}{10\% labeled} \\
\cmidrule(lr){3-4}\cmidrule(lr){5-6}\cmidrule(lr){7-8}
 &  & DSC & Sens & DSC & Sens & DSC & Sens \\
\midrule
Mean Teacher \textcolor{gray}{(NeurISP'17)}   & T2--T1CE & 0.4579 & 0.4226 & 0.6269 & 0.5915 & 0.6946 & 0.6412 \\
        & T1CE     & 0.1265 & 0.3526 & 0.0895 & 0.4502 & 0.0909 & 0.3972 \\
\hline
DTC \textcolor{gray}{(AAAI'21)}  & T2--T1CE & 0.4378 & 0.4296 & 0.6071 & 0.5550 & 0.6607 & 0.6034 \\
        & T1CE     & 0.1049 & 0.4036 & 0.1036 & 0.3518 & 0.1159 & 0.4014 \\
\hline
DTML \textcolor{gray}{(PRCV'21)} & T2--T1CE & 0.4455 & 0.4173 & 0.6033 & 0.5482 & 0.6777 & 0.6346 \\
        & T1CE     & 0.0628 & 0.3840 & 0.0860 & 0.3137 & 0.0943 & 0.4341 \\
\hline
SASS \textcolor{gray}{(MICCAI'20)} & T2--T1CE & 0.4119 & 0.3733 & 0.6187 & 0.5708 & 0.6814 & 0.6293 \\
        & T1CE     & 0.1175 & 0.3140 & 0.1059 & 0.2951 & 0.1135 & 0.1135 \\
\hline
UAMT \textcolor{gray}{(MICCAI'19)} & T2--T1CE & 0.4601 & 0.4368 & 0.6445 & 0.6181 & 0.7069 & 0.6873 \\
        & T1CE     & 0.1209 & 0.3881 & 0.0834 & 0.4129 & 0.1066 & 0.4793 \\
\hline
UMCT \textcolor{gray}{(MedIA'20)} & T2--T1CE & 0.4629 & 0.4485 & 0.6440 & 0.6018 & 0.7187 & 0.6590 \\
        & T1CE     & 0.0643 & 0.4382 & 0.1122 & 0.4067 & 0.1636 & 0.4284 \\
\hline
SPCT \textcolor{gray}{(MedIA'21} & T2--T1CE & 0.4817 & 0.4257 & 0.6478 & 0.5894 & 0.7191 & 0.6614 \\
        & T1CE     & 0.1219 & 0.3612 & 0.1213 & 0.3587 & 0.1307 & 0.4154 \\
\hline
DAFNet \textcolor{gray}{(T-MI'20)}  & T2--T1CE & 0.4960 & 0.4790 & 0.6120 & 0.5120 & 0.6830 & 0.6200 \\
        & T1CE     & 0.3940 & 0.3770 & 0.4980 & 0.3810 & 0.5390 & 0.4330 \\
\hline
CML \textcolor{gray}{(MedIA'23)}     & T1CE     & 0.4316 & 0.4718 & 0.6121 & 0.6403 & 0.6656 & 0.6917 \\
\hline
Ours    & T2--T1CE & \textbf{0.4989}& \textbf{0.5100} & \textbf{0.6359} & \textbf{0.6929} & \textbf{0.7203} & \textbf{0.7282} \\
\bottomrule
\end{tabular}}
\end{table*}
\paragraph{Quantitative Result}




Table \ref{tab:t1flair} and \ref{tab:t2t1ce} summarize performance for our proposed method and several multi-modal semi-supervised baselines on BraTs under 1\%, 5\%, and 10\% scenarios of labeled data. We only use two combinations of BraTs' modalities (e.g., T1, T1CE, T2, and FLAIR), such as T2-T1CE and T1-FLAIR, in our experiments. We select the combination based on recent empirical analyses on MRI modality in glioma segmentation \cite{support}. Additionally, the table illustrates both multimodal inference (using both input modalities in the inference stage) and monomodal inference (using only one modality) in order to point out the effectiveness of multimodal learning. For instance, the multi-modality achieves a huge gap seen in Mean Teacher with T2-T1CE at 0.4579 at the dice score, while single-modality only records 0.1265 when solely using T1CE. All of compared methods are multi-modal version by plugging our proposed modules into two distinct models and training by our semi-superivised cross-modality strategy.s

Notably, the CML method shows the state-of-the-art method while only using single-modality in the inference stage compared to the previous methods. The framework can be trained with multiple modalities, but only one is used in the testing, which unexpectedly restricts the performance when multiple modalities are available. Although it allows flexibility when some modalities are missing, complementary information from other channels cannot be exploited during inference. In contrast, our method solely focuses on leveraging all available modalities at both training and inference, resulting in an outperforming result. 

For T1-FLAIR shown in Table \ref{tab:t1flair}, our method substantially outperforms all compared methods by a large margin in both dice and sensitivity. Specially, at 1\% labeled data, it records a dice score of \textbf{0.7232} vs. CML \cite{cml} with 0.4337 (about \textbf{28\%} gain), and our method also improves about 8\% in sensitivity score (\textbf{0.6699} vs. 0.5900). A similar pattern can be seen in the scenario of 5\% and 10\%, where it reaches almost and in some cases exceeds in both dice and sensitivity. In terms of T2-T1CE in Table \ref{tab:t2t1ce}, our proposed method achieves state-of-the-art results in three different scales of labeled data, where roughly \textbf{7\%, 2\%, 6\%} improvement in dice score of 1\%, 5\%, 10\% scale of labeled data, respectively. Furthermore, we also report consistent improvement in sensitivity compared to most methods. 

\begin{figure*}[!h]
    \centering
    \includegraphics[width=\linewidth]{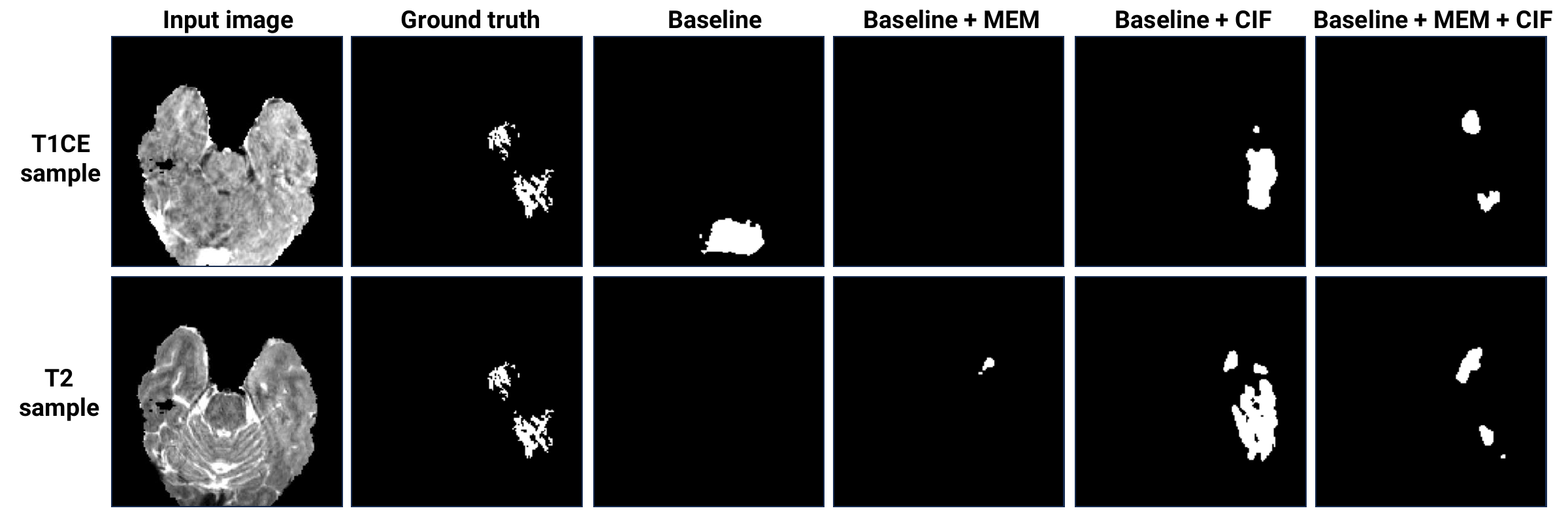}
    \caption{Visualization of the ablation study. The result stems from a model trained on T2-T1CE modality combination with 10\% of labeled data.}
    \label{fig:vis_ablation}
\end{figure*}

\paragraph{Qualitative Result}
The Figure \ref{fig:vis} illustrates the performance of our method compared to the strongest baseline named CML \cite{cml}, we report the prediction from two models trained on 10\% of labeled data with T2-T1CE and T1-FLAIR combination in BraTs dataset. It is clear that our method significantly improves in both big tumours and small tumours, which showcases the improvement in the higher precision of predicted pixels and the recall rate. For instance, at the first quarter (top left) of the Figure \ref{fig:vis}, it demonstrates the performance of a model trained with a combination of T2 and T1CE modality, where our proposed modules have the ability to sufficiently capture the tumor area while CML merely recognized a small of that or even cannot predict any tumor from the input image. The same behavior can be seen in the second quarter (top right), although CML method unexpectedly misses a complex structure of the tumor, ours is capable of partly marking it out. Additionally, the precision of predicted pixels is shown in the two bottom quarters (bottom left and bottom right); for example, the concave predicted by ours exactly matches the ground truth, while CML shows more false positive pixels as a dense area. 

\paragraph{Ablation study}
To verify the effectiveness of our proposed modules, we conducted rigorous evaluations on the dataset and compared the results in Table \ref{tab:ablation}, where Baseline presents the simplest architecture that only includes encoders and decoders for both modalities. We then incorporate each proposed component to investigate its individual contribution to the overall performance of the model. In addition, we also demonstrate in Figure \ref{fig:vis_ablation} the consistent improvement of each module by showing the performance on an example from a model trained on T2-T1CE combination on 10\% of labeled data.



\begin{table}[h]
    \centering
    \begin{tabular}{ccc|cccc}
        \toprule
        Baseline & MEM & CIF & Dice & Sens \\
        \midrule
        \checkmark &  &  & 0.6076 & 0.6012 \\
        \checkmark & \checkmark &  & 0.6371 & 0.5637 \\
        \checkmark &  & \checkmark & 0.7031 & 0.7754 \\
        \checkmark & \checkmark & \checkmark & 0.7203 & 0.7282 \\
        \hline
    \end{tabular}
    \caption{Ablation study on key components of the proposed
method on BraTs dataset (T2-T1CE) dataset with 10\% labeled data. Results are reported in Dice and Sensitivity.}
    \label{tab:ablation}
\end{table}

We initially investigated the effectiveness of the Modality-specific Enhancing Module; our baseline recorded 0.6076 in dice score and 0.6012 in sensitivity. We first investigate MEM module for its effectiveness in enhancing modality-specific knowledge within a modality by incorporating the MEM. The model subsequently shows a moderate increase to 0.6371, while the sensitivity slightly decreases to 0.5637. This improvement demonstrates that MEM is capable of enhancing representations by emphasizing modality-specific information, but may reduce recall for small or ambiguous regions. 

Furthermore, when we incorporate the CIF instead of MEM, the model shows a substantial performance gain in both metrics, with a better dice score of 0.7031 and sensitivity rising to 0.7754. This notable improvement indicates that CIF plays a crucial role in encouraging cross-modal interaction, allowing the model to effectively exchange complementary information from different modalities.
Then, when both MEM and CIF are present, the model achieves the best performance, with a dice score of 0.7203 and a sensitivity of 0.7282, which is about a 12\% increase in dice score (0.6076 vs. 0.7203) and sensitivity (0.6012 vs. 0.7282) compared to the baseline. This result shows that each component effectively completes its mission well. While MEM enhances modality-specific information, CIF bridges two modalities by fusing the enhanced features. 

In terms of qualitative results of the ablation study, Figure \ref{fig:vis_ablation} compares prediction on the same slice from T1CE and T2 modality. Specifically, a model that consists only of an encoder and a decoder as a baseline cannot completely recognize the area of the tumor, which either bounds a wrong place or even does not predict any pixels as foreground. However, when the MEM is incorporated, its ability to enhance modality-specific features allows the model to locate the area of the tumor and partly mark it out in T2 modality. Remarkably, the effectiveness of the CIF is obviously witnessed when it is utilized; the model is now capable of predicting the boundary of an exact tumour area compared to ground truth, but with a low precision. Given the success of two proposed modules, the framework consists of both of them and has higher precision while removing false positive pixels, resulting in a close match with ground truth. This result confirms that our proposed modules genuinely operate as it is supposed to. 

\section{Conclusion}
We presented a semi-supervised multi-modal segmentation framework that (i) enhances modality-specific semantics through a channel-wise \emph{Modality-specific Enhancing Module} (MEM), (ii) learns to exchange complementary cues via an adaptively learnable \emph{Complementary Information Fusion} (CIF) layer, and (iii) regularizes unlabeled data using a simple yet effective cross-modal prediction consistency. On BraTS 2019 (HGG subset), our method achieves remarkable performance under scarce-label regimes across two clinically meaningful modality pairs (T1–FLAIR and T2–T1CE), with large margins over SSL baselines. An ablation study confirms that our proposed MEM and CIF are complementary: MEM enhances modality-specific feature quality, while CIF facilitates effective cross-modal information flow, together producing more accurate segmentations.

\bibliography{aaai2026}


\end{document}